\documentclass[10pt,twocolumn,letterpaper]{article}

\usepackage[review,applications]{wacv}      

\usepackage{graphicx}
\usepackage{amsmath}
\usepackage{amssymb}
\usepackage{booktabs}

\usepackage[pagebackref,breaklinks,colorlinks]{hyperref}

\usepackage[capitalize]{cleveref}
\crefname{section}{Sec.}{Secs.}
\Crefname{section}{Section}{Sections}
\Crefname{table}{Table}{Tables}
\crefname{table}{Tab.}{Tabs.}

\def\wacvPaperID{239} 
\def\confName{WACV}
\def\confYear{2024}

\begin{document}

\title{Noise Crystallization and Liquid Noise: Zero-shot Video Generation using Image Diffusion Models}

\author{Muhammad Haaris Khan\\
\and
Hadrien Reynaud\\\\
Imperial College London\\
\and
Bernhard Kainz\\
}
\maketitle

\begin{figure}
\centering
\includegraphics[width = 0.85\hsize]{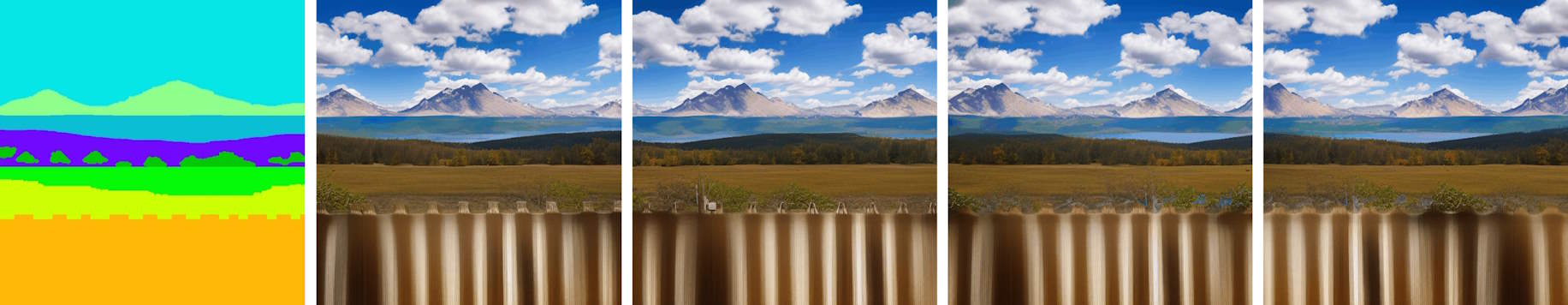}
\caption{Prompt-to-Video using noise crystallization (Segmentation map on left). Full animations viewable in supplementary materials \cite{supplementary}.}
\label{fig:dollyrecrystal2}
\end{figure}

\begin{figure}
\centering
\includegraphics[width = 0.85\hsize]{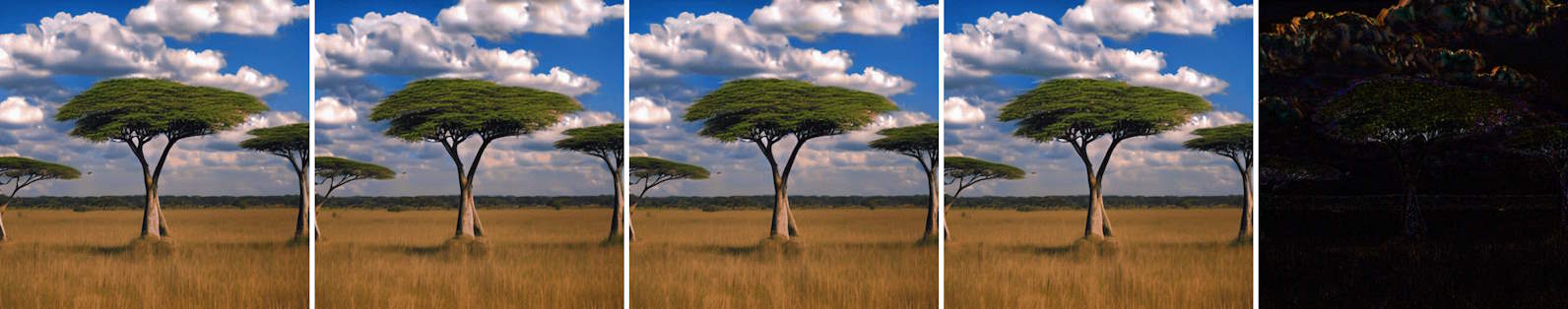}
\caption{Prompt-to-Video using liquid noise. Difference image (right) shows cloud motion and subtle swaying of grass and trees.}
\label{fig:noising_savannah}
\end{figure}

\begin{abstract}
   Although powerful for image generation, consistent and controllable video is a longstanding problem for diffusion models. Video models require extensive training and computational resources, leading to high costs and large environmental impacts. Moreover, video models currently offer limited control of the output motion. This paper introduces a novel approach to video generation by augmenting image diffusion models to create sequential animation frames while maintaining fine detail. These techniques can be applied to existing image models without training any video parameters (zero-shot) by altering the input noise in a latent diffusion model. Two complementary methods are presented. Noise crystallization ensures consistency but is limited to large movements due to reduced latent embedding sizes. Liquid noise trades consistency for greater flexibility without resolution limitations. The core concepts also allow other applications such as relighting, seamless upscaling, and improved video style transfer. Furthermore, an exploration of the VAE embedding used for latent diffusion models is performed, resulting in interesting theoretical insights such as a method for human-interpretable latent spaces.
\end{abstract}


\begin{figure}
\centering
\includegraphics[width = 1\hsize]{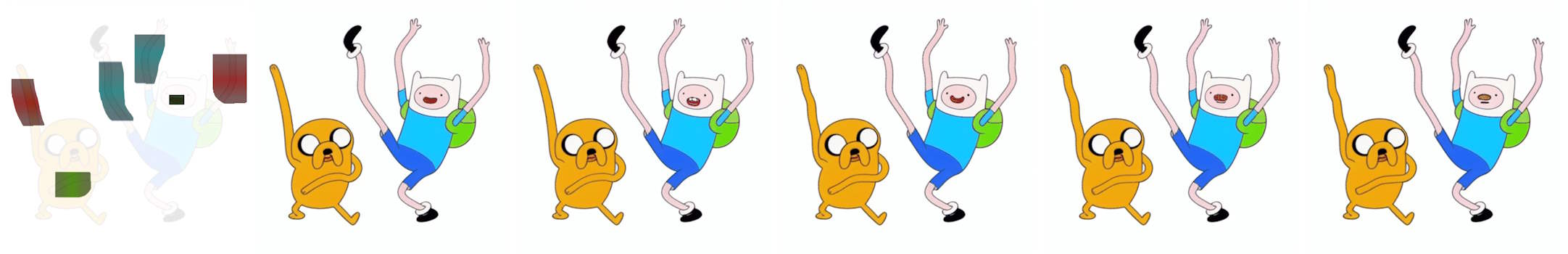}
\caption{Image-to-Video. A flow map used to move the arms and mouth of characters from \textit{Adventure Time} \cite{adventure_time}. Use permitted by \textit{Exceptions to copyright: Non-commercial research} \cite{gov_copyright}.
}
\label{fig:finn_jake}
\end{figure}

\begin{figure}
\centering
\includegraphics[width = 0.9\hsize]{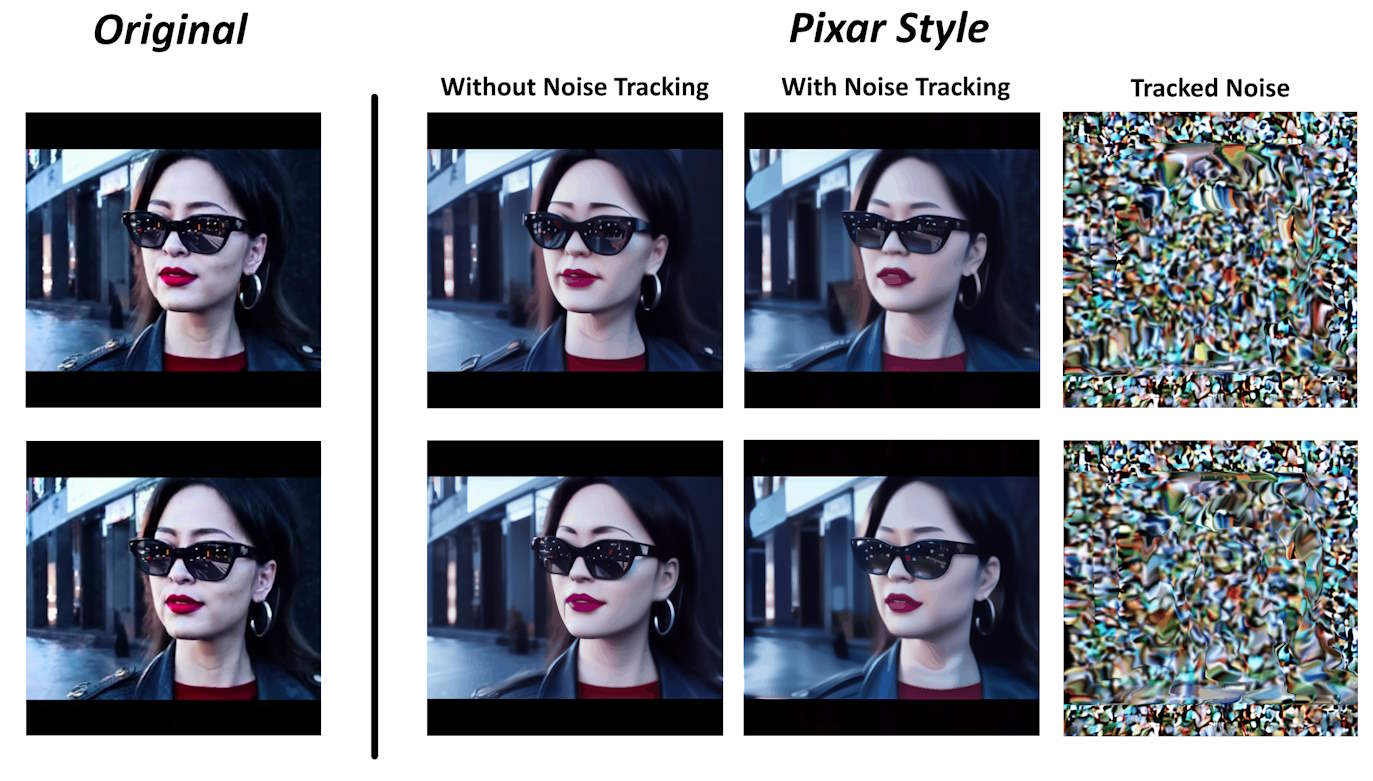}
\caption{Improved Video-to-Video style transfer. Notice the facial distortion in the examples without noise tracking (second column). Original video taken from \cite{videoworldsimulators2024}.}
\label{fig:vid2vid}
\end{figure}

\section{Introduction}
\label{sec:intro}

In recent years, \textit{diffusion models} have exploded in popularity.
Much of the progress has come in the form of image diffusion models, however they have often been compared to a ``slot machine'' experience \cite{fable2023showrunner}, due to the high unpredictability of consecutive generations. This inherent property is not suitable for long-form media such as video.

Some models are being trained specifically to produce videos that maintain temporal consistency. Examples include SORA from OpenAI \cite{videoworldsimulators2024} and Gen-3 from Runway \cite{gen3}. While the outputs of these models can be impressive, their training costs are prohibitively high, and the results remain difficult to control. \textit{Shy Kids}, a production company who used SORA to make \textit{Air Head}, said they generated around 300 videos to get a single decent shot for the final cut. Moreover, it took up to 20 minutes to generate a few seconds of video \cite{shykids2024}. Such high-compute video models may also pose environmental challenges in the long run.

\noindent\textbf{Contributions:} Our work proposes a new approach to video generation, which is to augment image diffusion models such that subsequent generations are consecutive frames of an animation, preserving fine details. The core concepts can be implemented on various pre-existing image diffusion models without any additional training (zero-shot), and thus without incurring the cost of training a video model.

The two methods presented in this paper are based on transforming the input noise of a latent diffusion model. Our \textbf{noise crystallization} technique (suitable for simpler motion) has highly consistent results and does not produce any artifacts. On the other hand, our \textbf{liquid noise} technique results in motion with a higher degree of freedom without resolution constraints.

Additional applications of the core concept include image-to-video, relighting, video style transfer, and seamless upscaling. This paper also includes an investigation into the nature of the embedding space of variational auto-encoders (VAE) used for latent diffusion models. We illustrate that these embedding spaces have a more disentangled latent space than previously thought and demonstrate that the VAE's tasks of colour conversions and upscaling are nearly independent.


\section{Related Work}
\label{sec:relatedwork}

Diffusion models have evolved significantly, recently achieving high-resolution image synthesis. Initially described in 2015 through an analogy to non-equilibrium thermodynamics, the problem was framed as numerically solving ODEs of particles undergoing Brownian motion. This was done using a thousand-step Markov chain to gradually convert one distribution into another \cite{sohldickstein2015deep}. 

The most prominent type of diffusion models, the \textit{Denoising Diffusion Probabilistic Model} (DDPM), was introduced in 2020 \cite{ho2020denoising}. 
Subsequent advancements include the \textit{Denoising Diffusion Implicit Model} (DDIM), which reduces the number of steps from one thousand to less than one hundred by employing a deterministic, non-Markovian process \cite{song2022denoising}. Unlike DDPMs, DDIMs incorporate noise solely from the model output. This is important for our paper as we focus on manipulating the noise directly.

\textit{Latent Diffusion Models} (LDM) were later popularised, performing diffusion in a perceptually compressed latent space, forming the basis for the \textit{Stable Diffusion} open-source line of models \cite{rombach2022highresolution}. This approach uses a U-Net architecture with multiple cross-attention blocks to enforce conditioning information from prompts. Additionally, self-attention within the U-Net enables both queries and keys to originate from the image itself.

Perceptual compression in LDMs is facilitated by a variational autoencoder (VAE), which maps images to distributions in the latent space, creating a smooth space suitable for sampling \cite{kingma2022autoencoding}. The default Stable Diffusion VAE (AutoencoderKL) minimizes four losses: reconstruction loss, KL divergence, perceptual loss (LPIPS), and adversarial loss \cite{AutoencoderKL}.

Being CNNs, U-Nets (such as in LDMs) have properties which are crucial for this work. Namely, translation equivariance, where spatial shifts in input result in corresponding shifts in output \cite{khetan2021implicit}.

Recently, \textit{ControlNet} emerged as an image conditioning method for ensuring high-level structural consistency in LDMs \cite{zhang2023adding}. ControlNet segmentation maps will be utilized in this work to maintain large-scale structure between frames, allowing the focus to remain on fine detail.


\section{Noise Crystallization}
\label{sec:method1}

The key idea of this technique is spatially translating the input noise of a latent diffusion model. The goal is to keep the same objects in the scene but in different positions, whilst preserving fine details. We base our experiments on Stable Diffusion v1.5 \cite{rombach2022highresolution}, and the sampler is DDIM \cite{song2022denoising}. In theory, a variety of open source models could work with this method.

\subsection{Prompt-Noise Decomposition}

The simplest version of this method involves creating a zero-shot panning effect by translating the input latent noise for a diffusion model laterally, with wrap-around. Several frames are generated and stacked to create a moving image sequence. This idea is inspired by the equivariant property of CNNs. If the diffusion model was near-linear, the \textit{homogeneity} property would allow the translation of the output to reflect the translation of the input.

\begin{figure}
\centering
\includegraphics[width = 0.9\hsize]{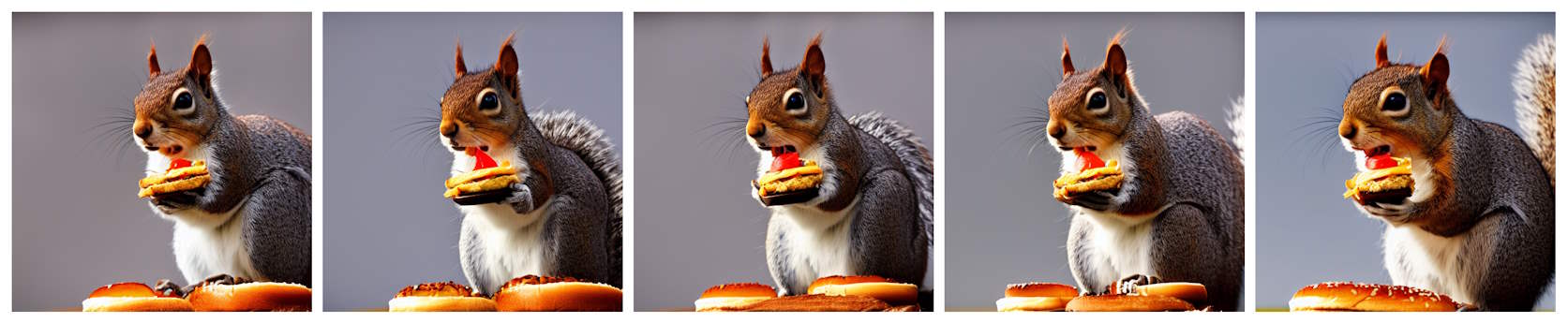}
\caption{Result of translating input noise to the left for each sample (no segmentation map usage).}
\label{fig:roll_attn}
\end{figure}

We find from Figure \ref{fig:roll_attn} that the above hypothesis is supported, as the output moves when the input noise is translated (ControlNet~\cite{zhang2023adding} is not being used here, so the outlines are less consistent). Our experiments show the attention maps (for each token in the prompt) is also translated alongside the noise. This introduces the idea that the combination of the noise and the prompt will position the attention map. Thus we hypothesize that any prompt-to-image generation can be characterised by the two inputs separately:\\
\begin{enumerate}
    \item The prompt, which determines \textbf{what} to display.
    \item The noise, which determines \textbf{where} to display it.
\end{enumerate}
In addition, these can be considered independent, since one can change where to centre the image by rolling the noise without altering what is displayed. The converse may or may not be true, as it cannot really be tested.

\subsection{Noise as a Crystal}

The following experiments will involve ControlNet segmentation maps \cite{zhang2023adding} to ameliorate large scale inconsistencies (such as outlines) so that the focus is on preserving fine details. It is important to apply the same transformations to the segmentation map as one did for the input noise for the animations to work.

It is convenient to think of the motion of the initial $64 \times 64$ noise latent as being confined to a 2D lattice or grid, similar to a crystal. In crystallography and metallurgy, a dislocation is an imperfection in the crystal where some atoms have \textit{glided/slipped}, and then snapped to a new location in the lattice \cite{pham2023microscopic}.

Due to the low resolution of the latent ($64 \times 64$), it is tempting to interpolate between array values. In actuality, the pixel values themselves need to stay the same for consistency. CNNs exhibit equivariance, but this property is compromised when pixel values change due to interpolation. Interpolation introduces horizontal motion blur in the noise, which blurs the final image. Post-interpolation noise is no longer white; high-frequency reduction and cross-correlations between neighbouring pixels introduce a pink noise characteristic, violating i.i.d. assumptions. Re-normalizing noise does not negate this effect, and potentially renders the images unrecognizable.

\subsection{Prompt-to-Panning Effect}

\begin{figure}
\centering
\includegraphics[width = 1\hsize]{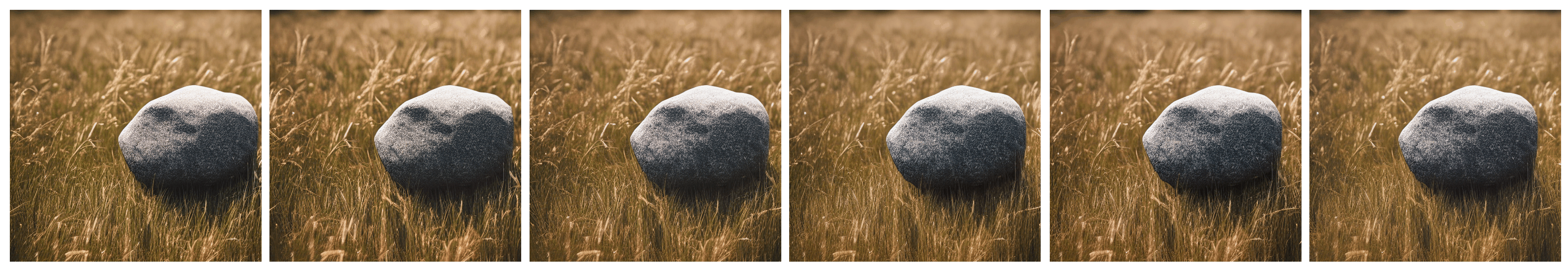}
\includegraphics[width = 0.4\hsize]{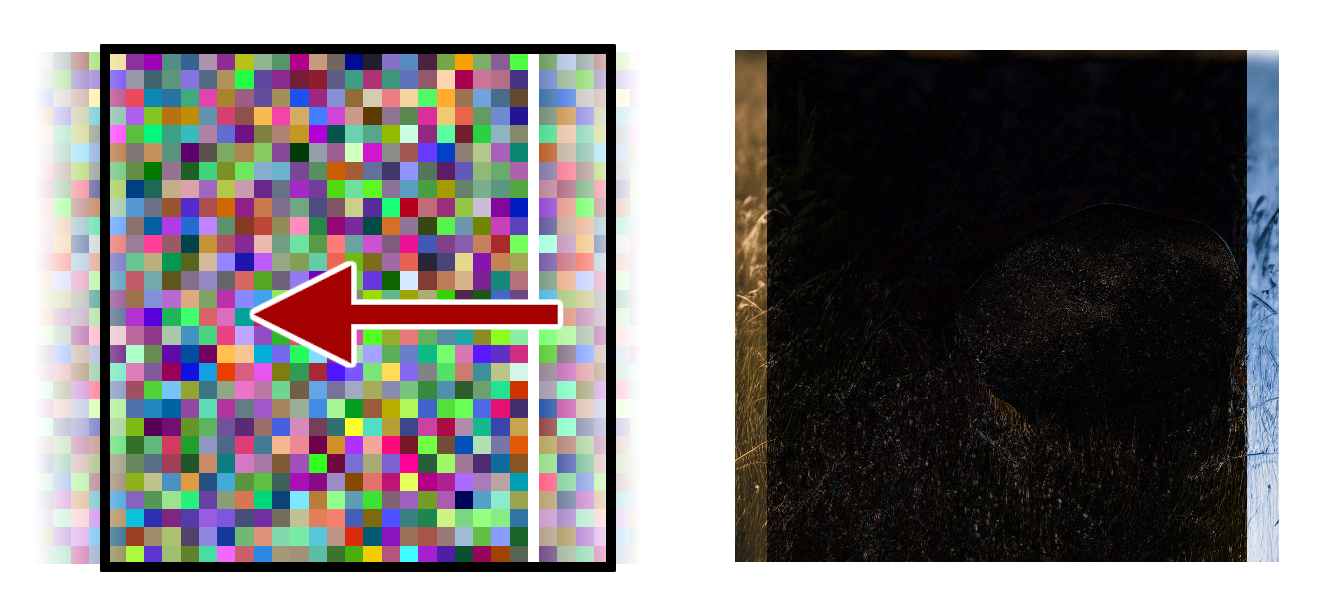}
\caption{Frames from the 2D panning effect, as well as the method, and difference image showing that the two images can be quite similar in details.}
\label{fig:pan}
\end{figure}

With the help of ControlNet \cite{zhang2023adding}, we arrive at the final 2D panning effect (Figure \ref{fig:pan}). The result in itself is intriguing due to the fact that equivariance seems mostly preserved even though this is not a pure CNN architecture. The Stable Diffusion architecture also uses self-attention layers \cite{rombach2022highresolution}, whose training depends on position. Our experiments indicate that this primarily leads to lighting inconsistencies between frames.

Evidently, the determinism of the DDIM scheduler is essential. Consistent results are not reproducible with a stochastic scheduler such as DDPM, because of the novel, uncontrolled noise added at each reverse diffusion step.

\subsection{Prompt-to-Parallax Dolly Effect}

\begin{figure}
\centering
\includegraphics[width = 1\hsize]{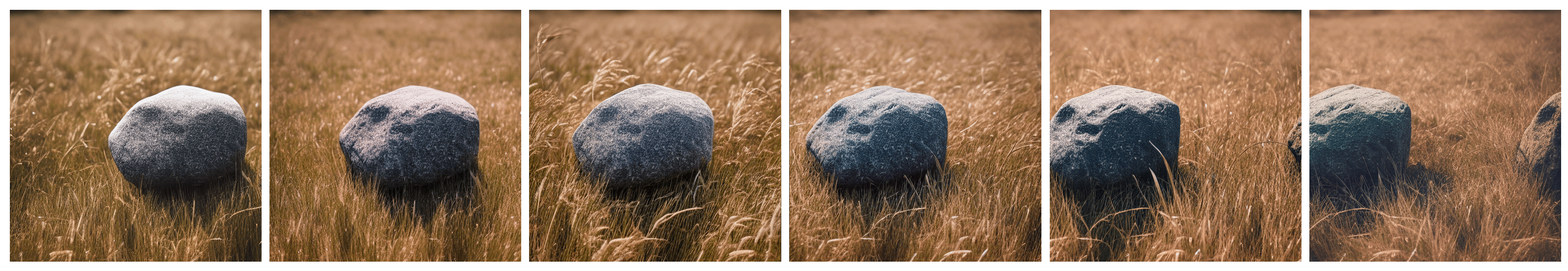}
\includegraphics[width = 0.6\hsize]{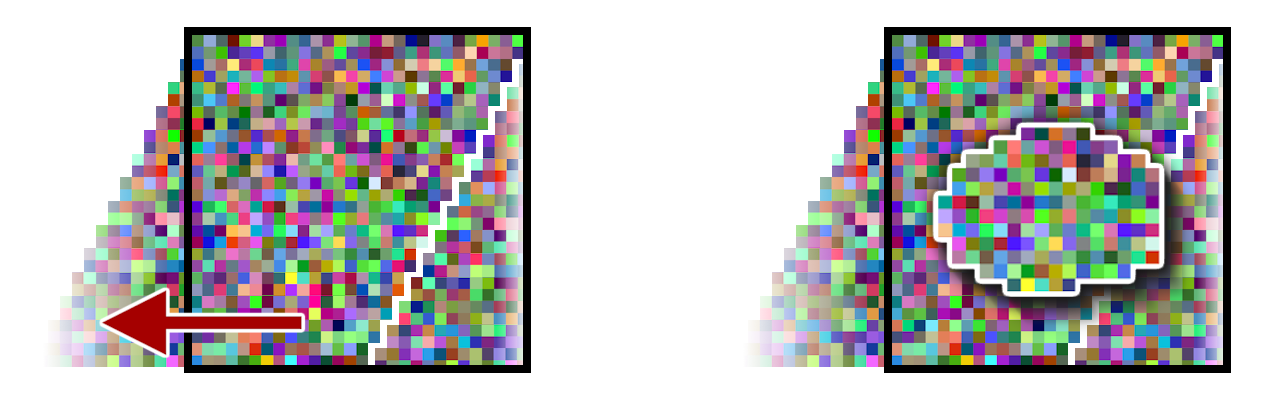}
\caption{Frames from the dolly parallax effect, as well as the method to glide the lattice and form the mosaic. Note the variation in the grass length and lighting.}
\label{fig:dolly}
\end{figure}

To create a parallax effect that appears to be a real camera movement, and not just an image pan, the foreground must move much faster than the horizon. An affine shear operation is calculated and discretised, which controls the gliding of each row. However, for complex scenes, a simple affine shear is not sufficient to be realistic. To avoid the skewing of near objects, one needs to take the original masked noise that belonged to the object, and paste it over the latent in the new object's position, creating a mosaic (Figure \ref{fig:dolly}).

\begin{figure}
\centering
\includegraphics[width = 0.9\hsize]{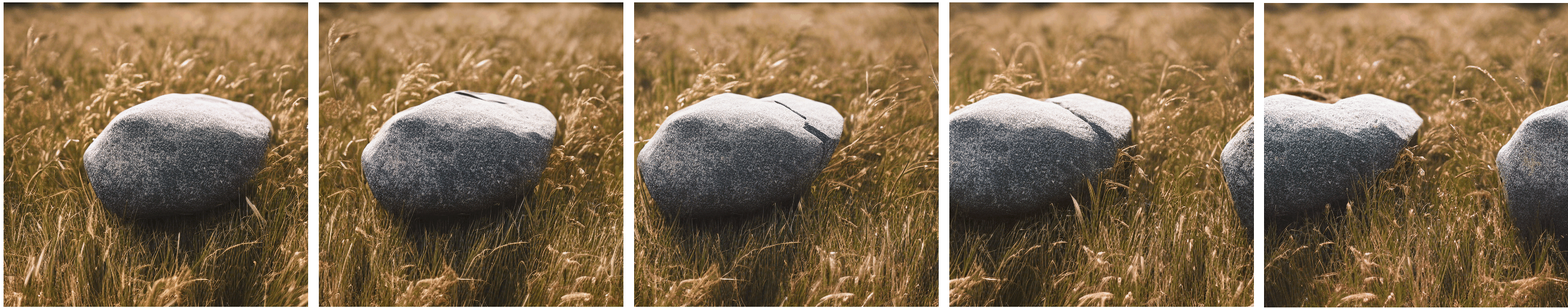}
\caption{Frames from dolly experiments where the noise and segmentation map is only changed 50\% of the way through the process. Notice the remnant of the old skewed rock now remains behind the new one, as it was partially generated. Figure \ref{fig:dollyrecrystal2} shows an example without near objects.
}
\label{fig:dollyrecrystal}
\end{figure}

\begin{figure}
\centering
\includegraphics[width = 1\hsize]{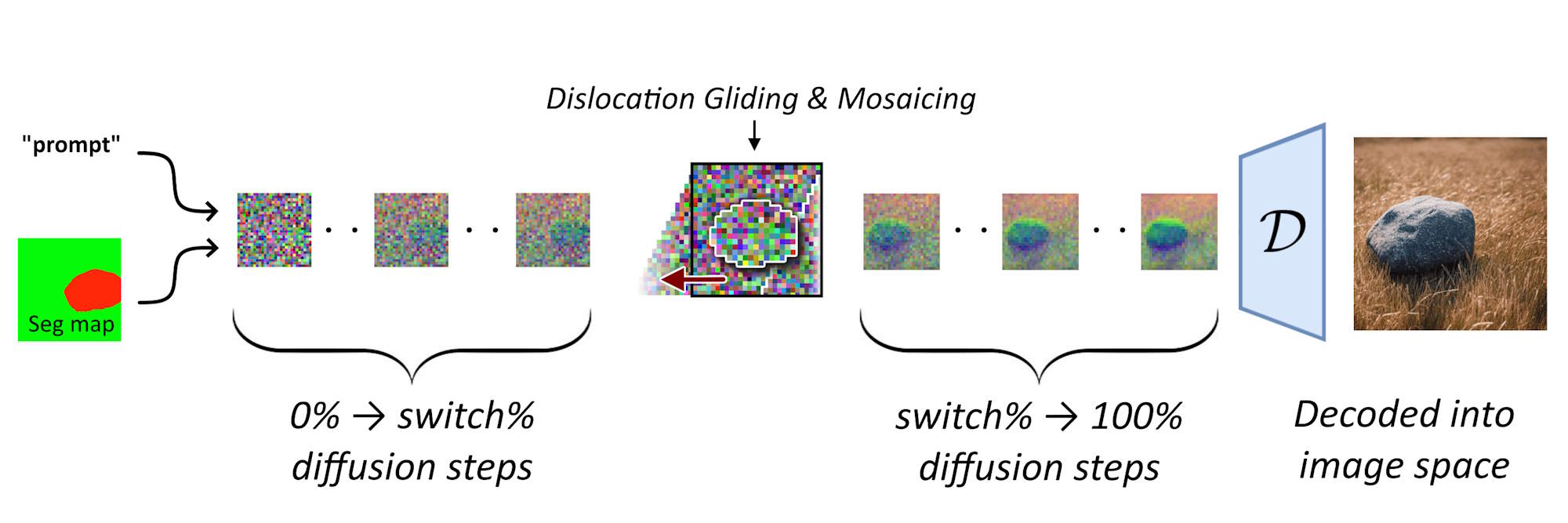}
\caption{Final diagram for the \textit{Noise Crystallization} method, which includes recrystallization at a certain switch percentage. In practice, the first diffusion steps can be reused for every frame.}
\label{fig:dollyrecrystaldiagram}
\end{figure}

\subsection{Recrystallization}

One of the issues in noise crystallization is the lack of consistency in long animations. We can use a heat treatment analogy to perform part of the diffusion process with the original noise, and then change to the transformed noise mosaic (and segmentation map) for the rest of the process. The motivation is that this helps control the lighting variations, as the initial phases of the diffusion process determine the large-scale structure. Once those are `set', we can \textit{recrystallize} at the \textit{switch \%} to generate the fine details in the new position (Figure \ref{fig:dollyrecrystal}). The best case scenario for this method is when there are no occluding objects (Figure \ref{fig:dollyrecrystal2}). We find that a gradual transformation throughout the process (annealing) does not work as well.


\section{Characterising the VAE}
\label{sec:vae}

Since traditional interpolation techniques amplify low frequencies, attention should be given to non-linear techniques instead. We leverage the VAE from Stable Diffusion as an upscaler that can hallucinate high frequency details, given its training.

\subsection{VAE Behavior}

This section deals with investigating the properties of the Stable Diffusion VAE (AutoencoderKL). Learning the VAE's true behaviour through experimentation is useful for determining how to use it as an upscaler.

\begin{figure}
\begin{center}
\includegraphics[width = 0.55\hsize]{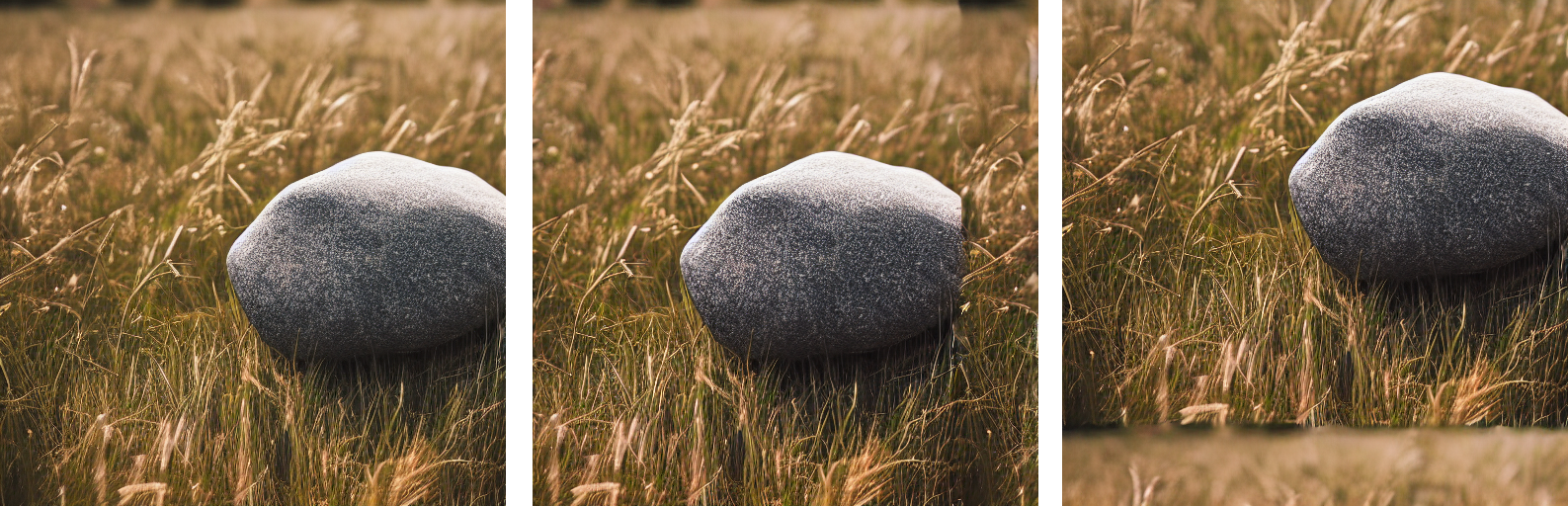}
\end{center}
\caption{Left: original image. Middle: image encoded, then rolled horizontally, then decoded. Right: image encoded, then rolled vertically, then decoded.}
\label{fig:vae_roll}
\end{figure}

We find that the VAE is impacted more by local rather than global image content. From Figure \ref{fig:vae_roll}, we can see that the VAE happily decodes the image even after rolling the latent space to produce a nonsensical scene. We argue this is aided by the patch-based training objective (LPIPS) which enforces `local realism' \cite{rombach2022highresolution}.

Idempotency is important for stability and predictability, and may also suggest a one-to-one mapping between latent space and image space. We hypothesise the VAE is idempotent after the first application, as we assume that the image may need to `snap' to the nearest viable latent.

\begin{figure}
\begin{center}
\includegraphics[width = 0.9\hsize]{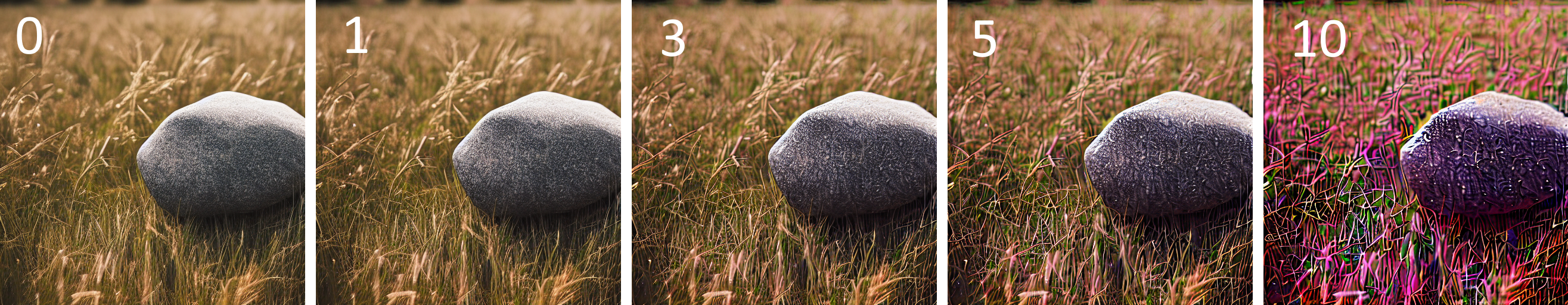}
\end{center}
\caption{Repeated application of the VAE, labelled with how many times the image was encoded and decoded. The leftmost image is the original.}
\label{fig:idemp}
\end{figure}

We observe from Figure \ref{fig:idemp} that the VAE is not idempotent as previously hypothesized due to veiny artifacts. The losses (other than reconstruction loss), encourage the VAE to take creative liberties while decoding. In addition, there is no large snapping behaviour on the first application for the natural images. This well-populated latent space is facilitated by the KL-divergence loss during training.

\begin{figure}
\centering
\includegraphics[width = 0.9\hsize]{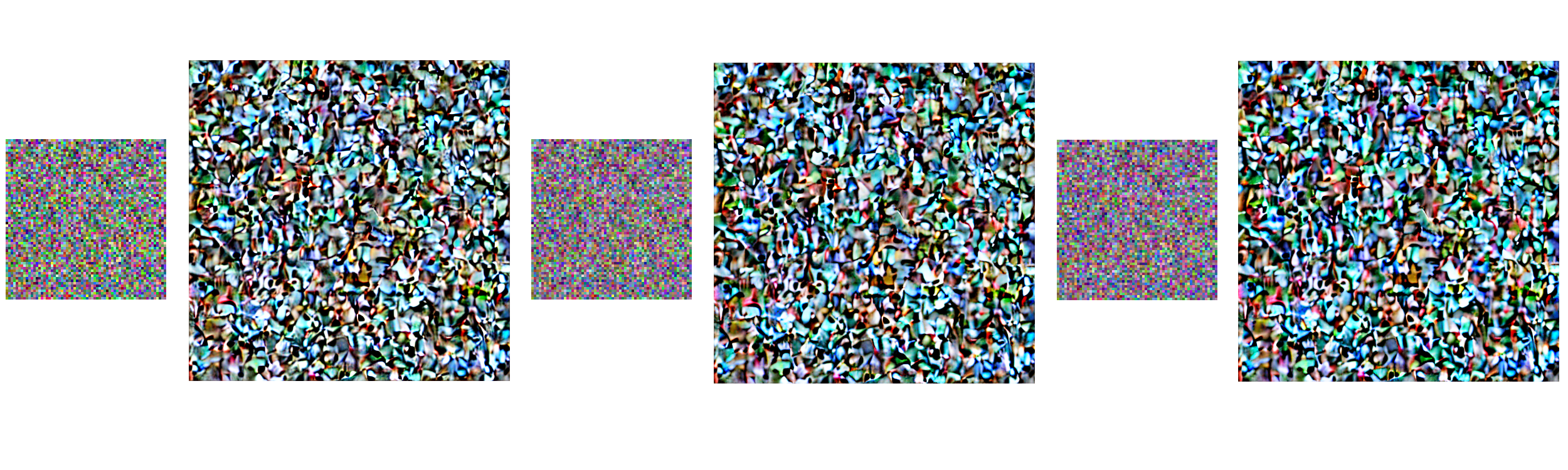}
\caption{Decoding and encoding a random noise latent with the VAE. The latent noises here are only represented by the first three channels.}
\label{fig:idemp_noise}
\end{figure}

For our purposes, it is important to know how the VAE handles noise, which is out-of-distribution as it is not present in the VAE training set. Interestingly, the VAE is able to decode latent noise in a semi-predictable way. From the colour change in Figure \ref{fig:idemp_noise}, we see the snapping effect is more evident on the first application, suggesting a less populated latent space. Another explanation is the fact that the set of all natural images is just a minuscule subset of all possible images, most of which are noise.

\subsection{Human-Interpretable Latent Space}

Despite the latent space being 48 times smaller than the image space, the reconstructions are quite close. A hypothesis can be made that for spatial shifts of less than 8 image pixels, the rolling of information occurs mainly between the channels rather than the spatial dimensions. Following an experiment about the SDXL latent space \cite{sdxllatent}, it may be useful to investigate the roles of the channels in the VAE used in Stable Diffusion v1.5.

\begin{figure}
\begin{center}
\includegraphics[width = 0.8\hsize]{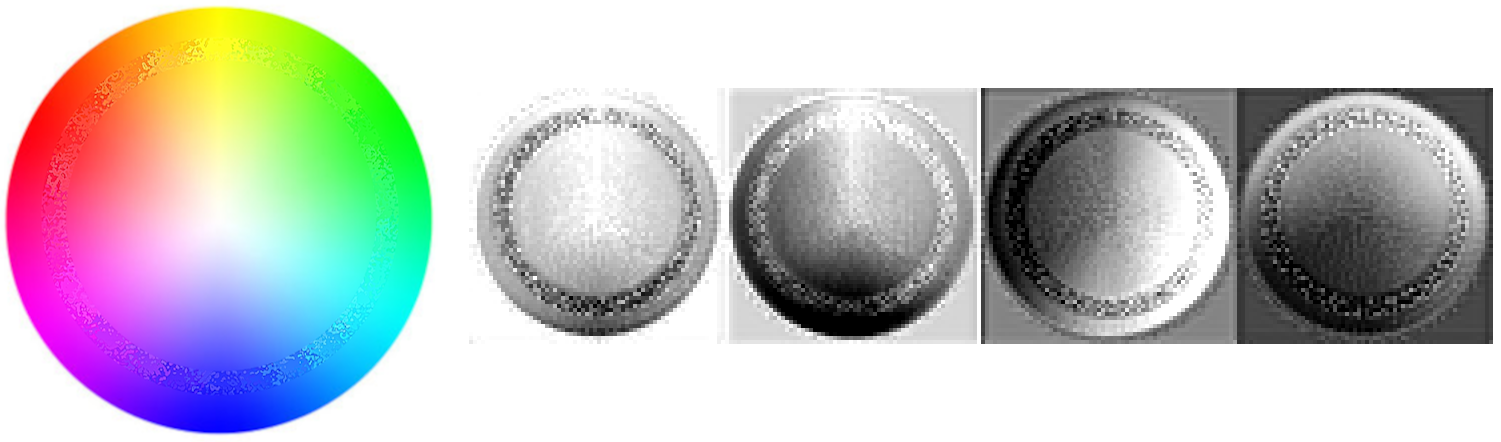}
\end{center}
\caption{Example colour wheel for characterising the channels of the SDv1.5 latent space. The wheel has a ring of texture in it to visualise if a single channel is responsible for details.}
\label{fig:wheel}
\end{figure}

From Figure \ref{fig:wheel}, it seems that there is no specific channel responsible for patterns or details. Moreover, we can surmise that the four channels are for luminosity, yellow, cyan, and magenta (inverted). The next step is to confirm by backpropagating a linear approximation of the VAE's latent-to-RGB behaviour (with no upscaling).

\begin{figure}
\centering
\includegraphics[width = 0.8\hsize]{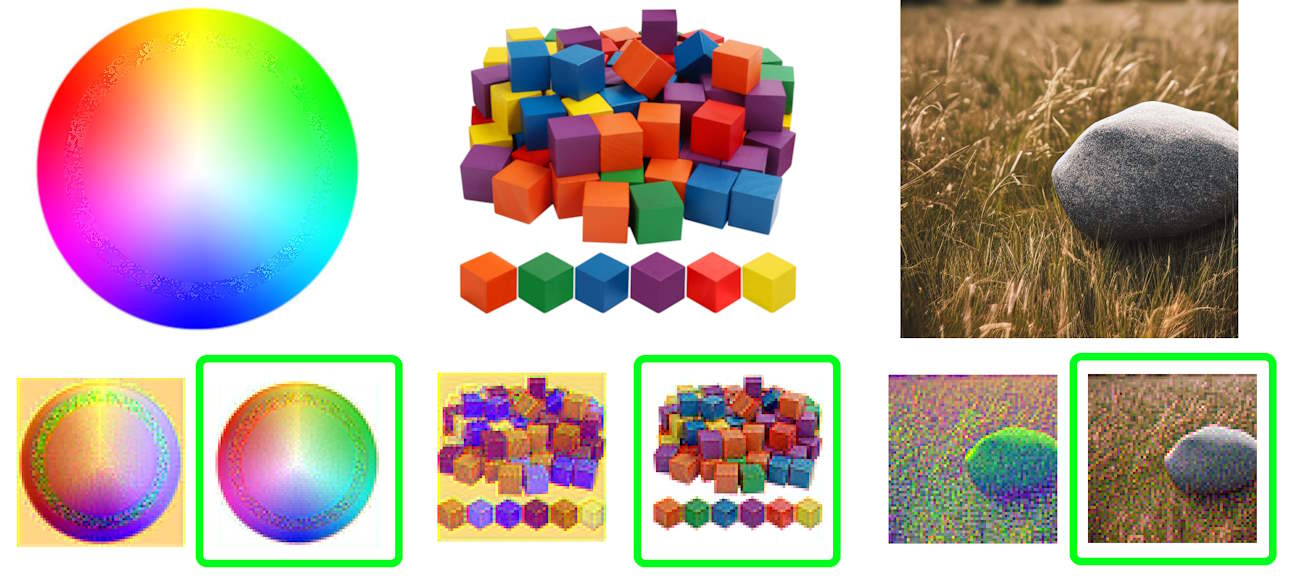}
\caption{Images next to their original latent visualisations (using 3 out of 4 channels), compared to the colour corrected RGB latents (green). Middle image from \cite{cubes}.}
\label{fig:colour_correct}
\end{figure}

\begin{align}
    \text{Weights} =
    \begin{bmatrix}
    43.89 & 16.35 & -35.44 & -21.61 \\
    29.65 & 44.57 & 32.10  & -29.15 \\
    36.27 & 5.53  &  28.26 & -82.53
    \end{bmatrix}
    \label{eq:matrix}
\end{align}
\begin{align}
    \text{Biases} =
    \begin{bmatrix}
        123.54 & 111.48 & 98.52
    \end{bmatrix}
    \label{eq:bias}
\end{align}

The regression (Eq.~\ref{eq:matrix} and Eq.~\ref{eq:bias}) does not expose any useless channel, which would have been identified by near-zero values. Thus, we conclude that none of the channels are specifically for details, and all four channels of the latent space contribute to colour encoding. Figure \ref{fig:colour_correct} shows a human-interpretable colour-corrected latent space.

Because we can approximate the colour conversion behaviour of the VAE using only two matrices, we conclude that most of the decoding work the VAE does is in image upscaling. In addition, the colour dimensions and spatial dimensions seem independent, resulting in a disentangled latent space. This suggests that, going forward, we can treat the VAE as a spatial upscaler, without being concerned about its behaviour with colour conversion.


\section{Liquid Noise}
\label{sec:method2}

Utilising the VAE to decode the latent at intermediate steps of the reverse diffusion process allows for movements smaller than one latent pixel. Thus, we suggest the liquid noise method. We treat the upscaled noise as freely deformable, as opposed to fixed on an array.


A critical issue is the VAE's alteration of statistical properties in out-of-distribution examples, \emph{i.e.}, samples not fully denoised. This occurs because the latent space properties change continuously with diffusion timestep $t$, while the VAE is trained only at $t=0$ (fully denoised).

By observing Q-Q plots, the new distribution deviates from an ideal Gaussian by means of kurtosis, in addition to mean and standard deviation. Kurtosis can be adjusted with $\hat{x} = \text{sinh}(\text{arcsinh}(\delta \cdot x))$ \cite{arcsinh2009}. The effect of kurtosis is limited, and we can even forego this correction due to the difficulty of determining $\delta$. For mean and standard deviation, we measure the original statistics of the latent  (\textit{channel-wise}) and then match them after re-encoding.

\subsection{Flow Maps}

Due to the freedom afforded by manipulating the decoded latent in $512 \times 512$ image space, we can use optical flow maps to delineate motion. A custom optical flow parser was created which utilizes the unused brightness channel to control the period in periodic sinusoidal motion. This is done so a single flow map can define the motion in the entire animation. Note that some scenes benefit from disabling wrap-around.

\begin{figure}
\centering
\includegraphics[width = 1\hsize]{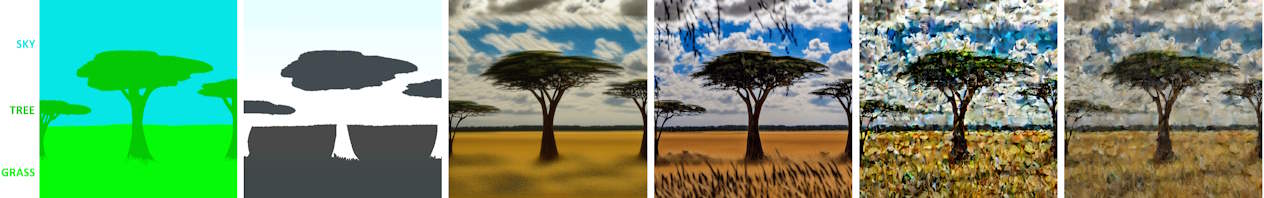}
\caption{From left to right: ControlNet segmentation map, flow map, gradient motion (with smearing), blocky motion (with artifacts). Decoded latents before and after variance reduction (switch = 70\%). The flow map has cyan for the sky (leftward motion), and grey-cyan for the trees and grass (horizontal swaying motion).}
\label{fig:flow_trees}
\end{figure}

\subsection{Artifact Treatment}

We observe in Figure \ref{fig:flow_trees} that if the flow maps have gradients, there will be smearing. If the flow maps have regions of constant colour, there will be artifacts, which need to be attenuated. We hypothesize that these artifacts come from the grid-like Voronoi pattern that the VAE produces when decoding a semi-diffused latent. When re-encoding the noised image after some movement, the receptive field of certain layers in the U-Net could be centred on these dark edges, leading to artifacts.

One method to reduce artifacts involves adding a small amount of noise (from a different seed) to the $64 \times 64$ latent after applying the transformation. Interestingly, introducing a small amount of noise to the decoded latent in the image space ($512 \times 512$) during the transformation process also proves beneficial (Figure \ref{fig:noising_savannah}). The added noise alters the final image by adding a small amount of graininess and lowering the overall contrast. This leads us to believe that lowering the contrast directly reduces the discernability of artifacts. Pushing this idea further, we find that simply reducing the variance of the latent before using the VAE also mitigates artifacts, without adding graininess. Investigating the decoded latent, we discover that lowering contrast directly prevents the dark edges of the Voronoi pattern to appear (Figure \ref{fig:flow_trees}). In the global pipeline, this process does not require further correction, since all data distributions are fixed afterward, as previously mentioned. This is stable so long as the standard deviation is re-adjusted before the mean.

\begin{figure}
\centering
\includegraphics[width = 1\hsize]{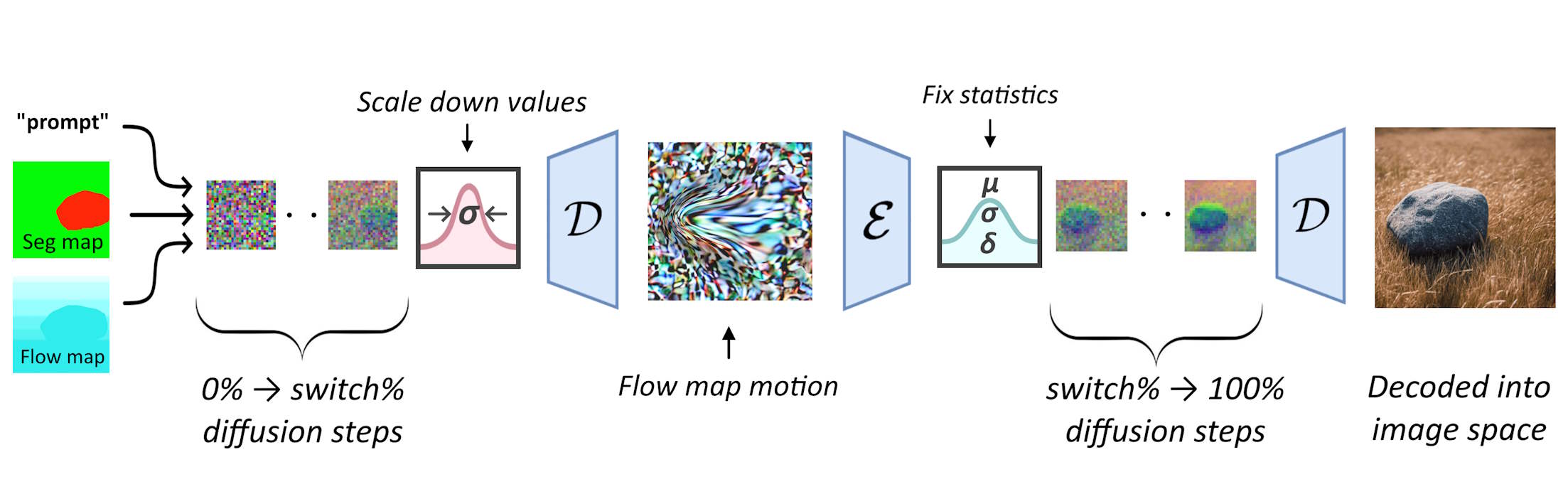}
\caption{Final diagram for the liquid noise method, with the help of the VAE and a custom flow map. The scale down factor is based on switch percentage. Compared to noise crystallization, high switch percentages need to be used.}
\label{fig:liquid_noise_diagram}
\end{figure}

We can conclude that when the artifacts are mitigated appropriately, liquid noise can produce impressive results for a zero shot video generation method. Through the mid-diffusion decoding and re-encoding of the latent, we even allow sub-pixel movements like rotating and zooming. Nonetheless, artifacts may remain in some samples, and this will be tied to the choice of initial noise seed. We note that this is not a major limitation given the speed of this approach compared to actual video generation models.


\section{Further Applications}
\label{sec:otherapplications}

We can also apply our core concept of manipulating the input noise of a diffusion model to other scenarios. We do not detail these approaches, as they are ancillary. We also skip the evaluation for certain methods, which we only describe here.

\subsection{Image-to-Video}

We can extend the image-to-image setup of diffusion (noising an image) to animate a pre-existing picture, leading to image-to-video. 
Segmentation maps are not needed, but flow maps are still used to perform the animation. As can be seen, this can save animators time, as the diffusion model produces consistent line thicknesses, precluding the need to redraw every frame (Figure \ref{fig:finn_jake}).

\begin{figure}
\centering
\includegraphics[width = 1\hsize]{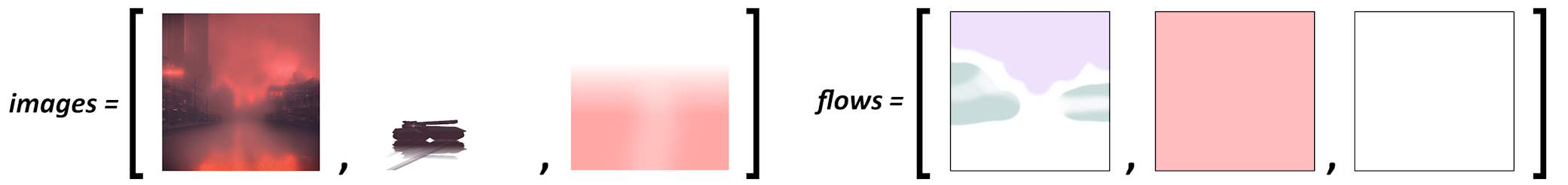}
\includegraphics[width = 1\hsize]{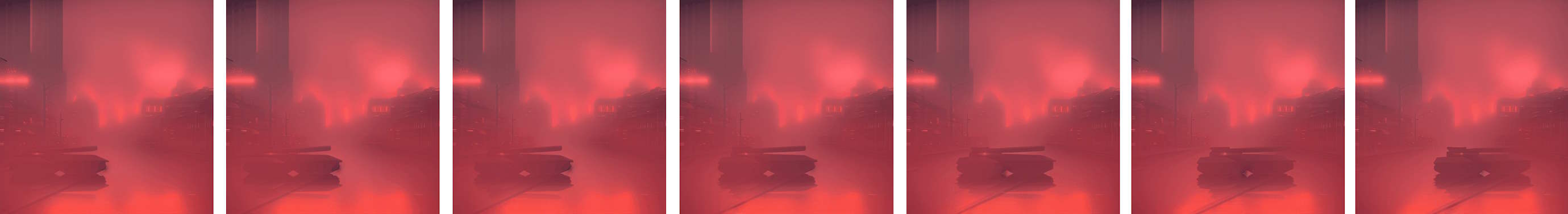}
\caption{Animating with layers. This method helps combine pre-existing images into animated scenes, using diffusion to add detail.}
\label{fig:layers_demo}
\end{figure}

Furthermore, if one passes in multiple images and corresponding flow maps, an animation with layers can be performed (Figure \ref{fig:layers_demo}). This solves a lot of previous problems with occlusion, and allows for more complex movements. Compared to traditional digital art, the concept of layers is something usually missing in generative AI methods. In this approach, the images are transformed and then alpha blended together. Several layers of the added noise are also tracked to reduce flickering. Instead of choosing layers from different images, one can also prepare layers via image inpainting of an existing image to reveal occluded areas for each layer.

\subsection{Object Permanence and Relighting}

\begin{figure}
\centering
\includegraphics[width = 0.4\hsize]{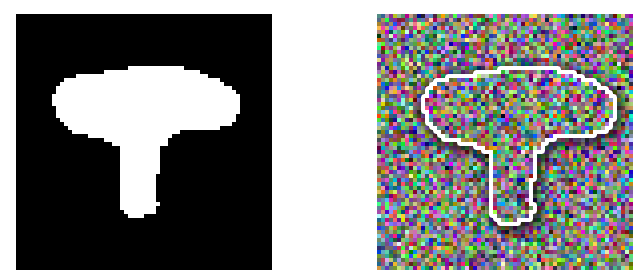}
\includegraphics[width = 1\hsize]{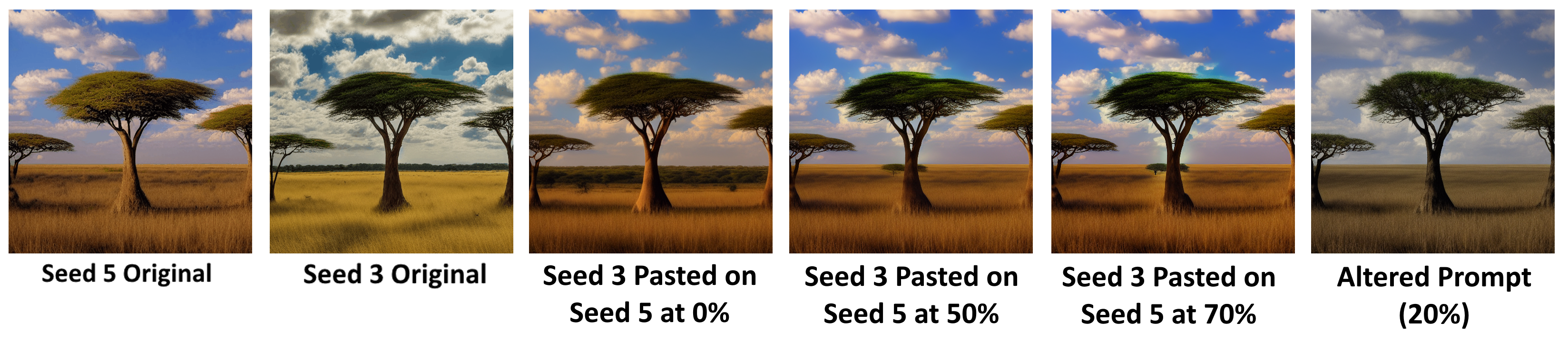}
\caption{Demonstration of object permanence by pasting a foreground noise shape from one seed onto background noise from a another seed. The percentage is how far each individual latent is diffused before they are combined together.}
\label{fig:relighting_demo}
\end{figure}

The concept of object permanence is to use masks to copy noise from one seed to another. As we have shown throughout this work, many properties from objects in diffused images come from the initial noise. Thus, compositing noise from different seeds allows re-use of the same unique object. We observe that the shapes of the foreground object are mostly preserved when pasted onto the background, but with different lighting. If one wants to maintain the same object lighting, it is possible to combine the latents after some diffusion steps (Figure \ref{fig:relighting_demo}). At high percentages, lighting halos appear if the mask is not exact. Note that without \textit{Null-text Inversion} \cite{mokady2022nulltext}, one cannot take a pre-existing image (such as a face) and fully find the source noise, making face relighting difficult.

\subsection{Video-to-Video with Noise Tracking}

Video-to-video style transfer involves applying a filter to each frame of a video, specifically using diffusion in our context. Inconsistencies will occur with diffusion due to the noise being fixed under the moving objects. However, temporal consistency can be enhanced if the added noise was deformed via an optical flow method to be tracked under the objects (Figure \ref{fig:vid2vid}). For this method, the custom parser was not used, as periodic motion was not needed. This improved style transfer method could potentially be used to render digital environments like games using diffusion (if leveraging a fast \textit{latent consistency model} \cite{luo2023latent}).

\subsection{Seamless Upscaling}

\begin{figure}
\centering
\includegraphics[width = 0.9\hsize]{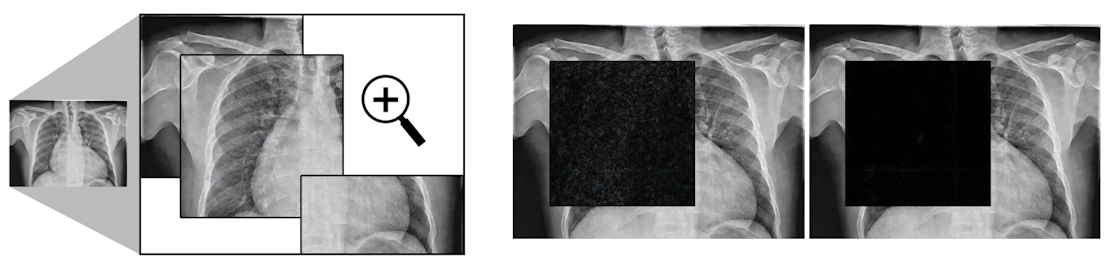}
\caption{Left: demonstration of the seamless upscale use case. Right: comparison of the standard approach vs the noise tracking approach. The black areas are difference images showing that the standard approach has discrepancies between upscales, unlike with noise tracking. X-ray adapted from \cite{radiopaedia}.}
\label{fig:magnifydemo}
\end{figure}

\begin{figure}
\centering
\includegraphics[width = 0.6\hsize]{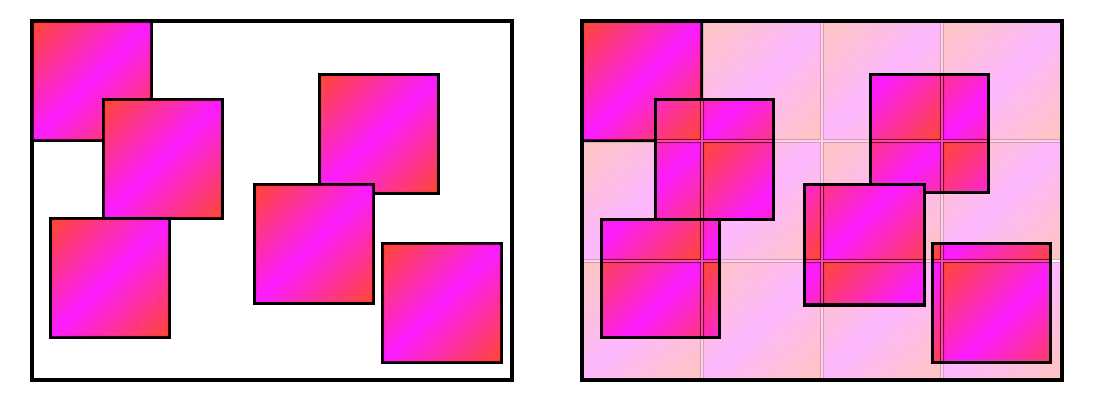}
\caption{Diagrams of different upscaling approaches. Left: standard approach, where the same noise is typically "stamped" for every region. Right: noise tracking approach, where the noise is sampled from a tiled background noise.}
\label{fig:seamless_upscale_diagram}
\end{figure}

The project's core concept of manipulating the input noise for consistency is also useful for stitching overlapping diffused images together. One example of this is piecewise upscaling. Typically, the seed or noise is not considered important, because of the default stochastic noise scheduler. However, for important applications such as medical imaging or document upscaling, one may want to minimise visible upscaling artifacts. To demonstrate this concept, we consider a scenario where a clinician uses a magnifying glass to closely examine a specific area of an X-ray (Figure \ref{fig:magnifydemo}). This can be interpreted as digitally enlarging the area to a resolution of $512 \times 512$. Since the doctor can zoom in on any arbitrary region, they may look at the same body part in two separate (overlapping) upscales of the original image. In this situation, consistency is crucial to maintain information across all upscaled tiles. With a random or frozen noise, disagreements show up between overlapping images, however if the noise also moves under the magnifying glass (Figure \ref{fig:seamless_upscale_diagram}), ideal results can be achieved with no dissimilarities. This concept could also be applied to \textit{inpainting} and \textit{outpainting}.

\section{Experiments}
\label{sec:experiments}

FVD score was considered for evaluation \cite{unterthiner2019accurate}; however, FVD measures visual quality, temporal coherence, and sample diversity \cite{unterthiner2019accurate}, while only the second property is relevant to compare approaches in this work. Visual quality and sample diversity change depending on the pre-existing image model that our methods run on top of.

\begin{figure}
\centering
\includegraphics[width = 1\hsize]{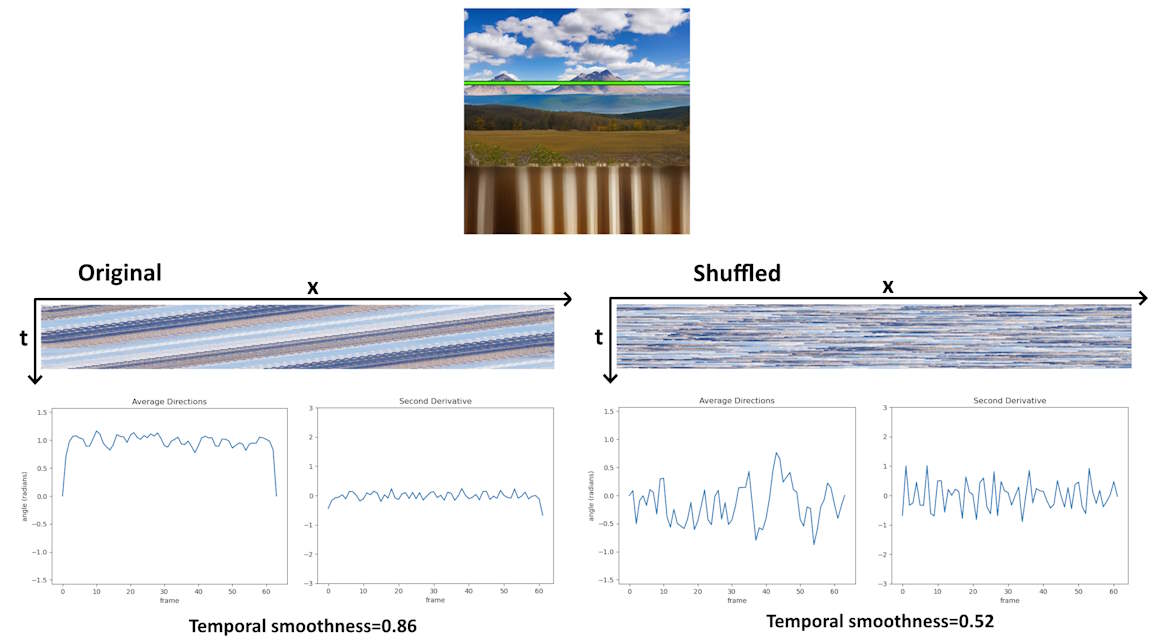}
\caption{Temporal smoothness comparison of original leftwards motion from one row (green) of the vista animation, and the same X-T slices but shuffled randomly.}
\label{fig:temporal_smoothness}
\end{figure}

A more promising way to evaluate temporal consistency is via \textit{spatio-temporal (X-T) slices} \cite{X-T2003}. In this method, we only take a single row or column of a frame, and observe its evolution over time (Figure \ref{fig:temporal_smoothness}).

We propose a new metric for numerically determining temporal smoothness from these X-T slices. Non-temporally consistent videos result in X-T graphs having edge directions that change more frequently over time. We start by finding the Sobel magnitudes $r$ and orientations $\theta$ of the X-T slices. Edges whose orientation is $<-\frac{\pi}{2}$ or $>\frac{\pi}{2}$ are discarded so the final vector always points forward in time. The polar angle $\theta$ of the weighted vector sum ($\sum r e^{i\theta}$) will be the overall direction of that slice, which is done for all slices. Next, the second derivative of overall directions over time is taken. The standard deviation of this gives a value to the roughness of the motion. For smoothness, we take $e^{-\text{roughness}}$, where the smoothest value achievable is 1.0.

We compare to: 1) \textit{Text2Video-Zero}: another zero-shot video method based on Stable Diffusion \cite{text2videozero}. Due to the paucity of zero-shot methods in literature, we also compare to: 2) \textit{AnimateDiff}: which aims to be a low-cost addition to Stable Diffusion~\cite{animatediff}; and 3) Open-Sora: A full video model~\cite{opensora}. Samples were chosen from the different methods, and the average temporal smoothness metric was calculated using five horizontal rows. Each X-T slice is limited to 16 frames, and all frames are cropped to have a size of $512\times512$.

Our methods shows high temporal consistency, appearing to be on par with trained competitors, and outperforming the zero-shot competitor (Table \ref{table:xt_results}). However, this doesn't take into account the simplicity of our animations. Our examples do not contain complex occlusions, and motion is enforced by traditional array indexing. In contrast, Open-Sora's full video model attempts more ambitious movements, such as 3D object rotations. AnimateDiff offers a middle ground, adding motion to Stable Diffusion with slightly less consistent results. Current zero-shot methods do not possess both flexibility and consistency, with Text2Video-Zero leaning towards the former, and our methods leaning towards the latter. Full video models generate realistic, complex motions with minimal user input but demand more training. Our zero-shot methods can produce temporally consistent, controllable, yet simplistic movements, requiring user-made flow maps. One should be wary of solely using consistency to judge these methods, as it gives limited insights on model capability, flexibility, or quality. Nonetheless, our aim here is to measure temporal consistency, and this was achieved. Table \ref{table:speed} shows our methods achieve faster inference speeds than the open-source competitors.

\textbf{\begin{figure}
\centering
\includegraphics[width = 1\hsize]{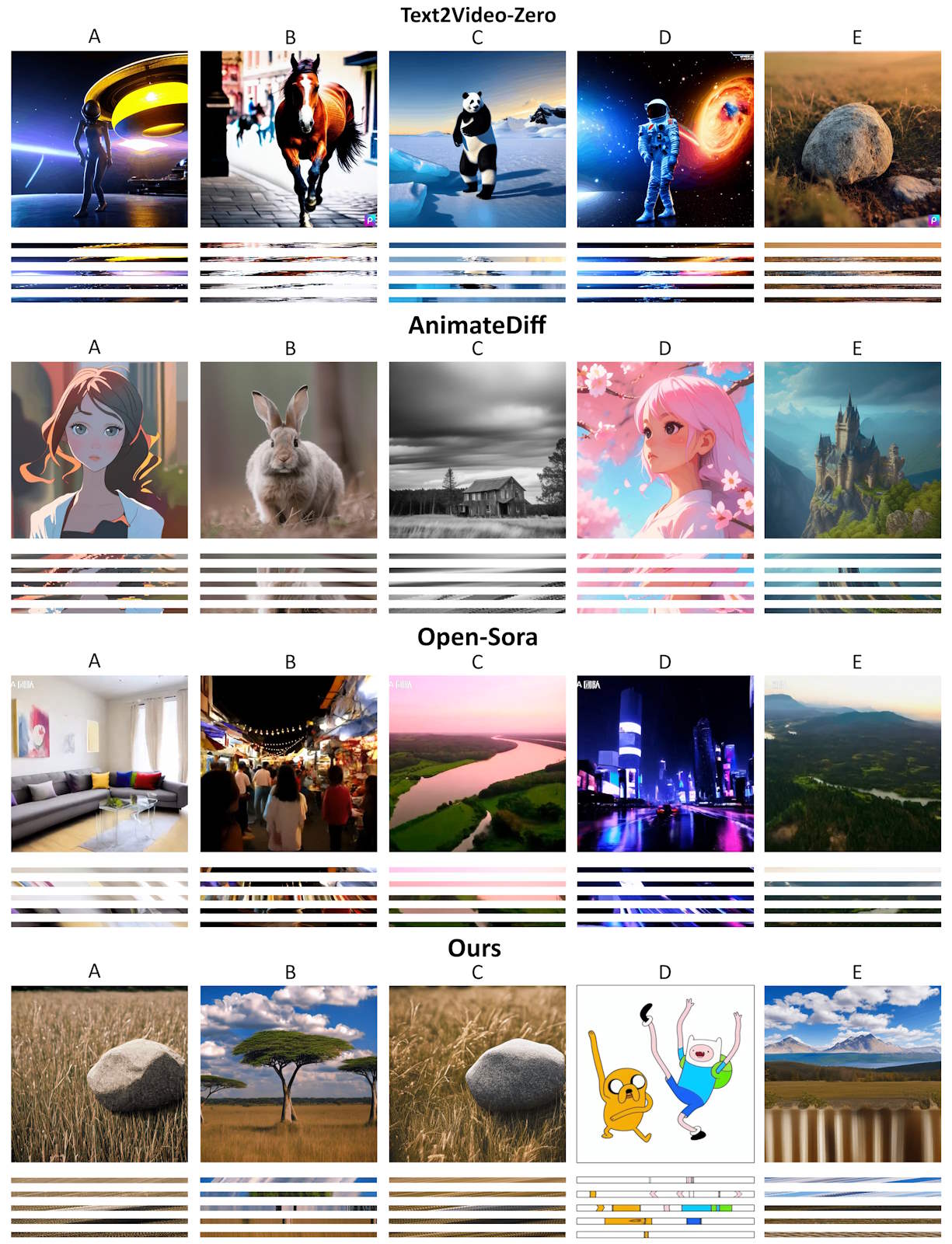}
\caption{Sample animations and five rows of corresponding X-T slices from Text2Video-Zero, AnimateDiff, Open-Sora, and our methods. Time on the y-axis. Taken from \cite{animatediff, opensora, text2videozero}. \textit{Adventure Time} image from \cite{adventure_time}.}
\label{fig:xt_results}
\end{figure}}

\begin{table}
\centering
\scriptsize
\begin{tabular}{|| c | c | c ||} 
    \hline
    \multicolumn{3}{|| c ||}{Text2Video-Zero \cite{text2videozero}}\\
    \hline
    Sample & Description & Temporal Smoothness \\
    \hline
    A & Pose controlled motion & 0.48\\ 
    B & Low frame rate horse galloping & 0.41\\
    C & Pose controlled motion, still background & 0.31\\
    D & Pose controlled motion, moving stars & 0.40\\
    E & Low frame rate, still background & 0.32\\
    \hline
    \hline
    \multicolumn{3}{|| c ||}{AnimateDiff \cite{animatediff}}\\
    \hline
    Sample & Description & Temporal Smoothness \\
    \hline
    A & Low frame rate, stylised anime motion & 0.69\\ 
    B & Slow orbital movement & 0.77\\
    C & Pan and time-lapse & 0.66\\
    D & High frame rate anime motion & 0.76\\
    E & Orbital drone shot & 0.80\\
    \hline
    \hline
    \multicolumn{3}{|| c ||}{Open-Sora \cite{opensora}}\\
    \hline
    Sample & Description & Temporal Smoothness \\
    \hline
    A & Handheld camera move & 0.82\\ 
    B & Camera moving through crowd & 0.87\\
    C & Aerial drone shot & 0.78\\
    D & Flashing lights and reflections & 0.79\\
    E & Aerial drone shot & 0.72\\
    \hline
    \hline
    \multicolumn{3}{|| c ||}{Liquid \& Crystal Noise (ours)}\\
    \hline
    Sample & Description & Temporal Smoothness \\
    \hline
    A & Parallax dolly motion & 0.82\\ 
    B & Still camera, cloud and foliage motion & 0.88\\
    C & Pure 2D pan (crystal method) & 0.84\\
    D & Still camera, only limbs moving & 0.94\\
    E & Speeding, pan motion (crystal method) & 0.63\\
    \hline
\end{tabular}
\caption{Table of results for Figure \ref{fig:xt_results}, including descriptions.}
\label{table:xt_results}
\end{table}

%

\begin{table}
\centering
\scriptsize
\begin{tabular}{|| c | c ||} 
    \hline
    Method & Time per frame (s)      \\
    \hline
    Text2Video-Zero \cite{text2videozero}  & 1.75    \\
    AnimateDiff \cite{animatediff}  & 4.13    \\
    OpenSora \cite{opensora}        & 1.13\\
    Crystal Noise (ours)         & \textbf{0.77}\\ 
    Liquid Noise (ours)             & \textbf{0.84}\\
    \hline
    Stable Diffusion \cite{rombach2022highresolution} & 3.0     \\ 
    \hline
\end{tabular}
\caption{Inference times comparison between various open-source methods for video generation. All evaluations are conducted with the same A100 GPU and with the same parameters. All videos are 16 frames long with a resolution of $512\times512$. 30 diffusion steps are taken. Our Crystallization and Liquid Noise results use a switch percentage of 70\%. }
\label{table:speed}
\end{table}


\section{Conclusion}
\label{sec:conclusion}

Our work aimed to assess the viability of achieving a consistent animation using only zero-shot methods on top of image generators. Given the results, the viability of at least basic movements exceeded our initial expectations. One could even consider the outcome counter-intuitive, especially given the current `compute is king' zeitgeist in literature. Zero-shot methods have potential in deep learning for helping us be more resourceful, as there is the possibility of provoking greater capabilities from currently existing technology. With limited compute resources, there is a trade-off between consistency and flexibility.
We argue that the optimal solution is a hybrid mixture of compute model (combining zero-shot, lightweight, and full video models), where compute is determined by the ambitiousness of the request.

{\small
\bibliographystyle{ieee_fullname}
\bibliography{references}

\begin{thebibliography}{10}\itemsep=-1pt

\bibitem{cubes}
ataacars.
\newblock Montessori kira cube blocks, 2024.

\bibitem{radiopaedia}
Patrick Baird-Fraser.
\newblock Round pneumonia, 2022.

\bibitem{videoworldsimulators2024}
Tim Brooks, Bill Peebles, Connor Holmes, Will DePue, Yufei Guo, Li Jing, David Schnurr, Joe Taylor, Troy Luhman, Eric Luhman, Clarence Ng, Ricky Wang, and Aditya Ramesh.
\newblock Video generation models as world simulators, 2024.

\bibitem{AutoencoderKL}
CompVis.
\newblock Autoencoderkl, 2022.

\bibitem{animatediff}
Yuwei Guo, Ceyuan Yang, Anyi Rao, Zhengyang Liang, Yaohui Wang, Yu Qiao, Maneesh Agrawala, Dahua Lin1, and Bo Dai.
\newblock Animatediff, 2024.

\bibitem{ho2020denoising}
Jonathan Ho, Ajay Jain, and Pieter Abbeel.
\newblock Denoising diffusion probabilistic models, 2020.

\bibitem{arcsinh2009}
M.C. Jones and Arthur Pewsey.
\newblock Sinh-arcsinh distributions, 2009.

\bibitem{text2videozero}
Levon Khachatryan, Andranik Movsisyan, Vahram Tadevosyan, Roberto Henschel, Zhangyang Wang, Shant Navasardyan, and Humphrey Shi.
\newblock Text2video-zero: Text-to-image diffusion models are zero-shot video generators, 2023.

\bibitem{khetan2021implicit}
Naman Khetan, Tushar Arora, Samee~Ur Rehman, and Deepak~K. Gupta.
\newblock Implicit equivariance in convolutional networks, 2021.

\bibitem{kingma2022autoencoding}
Diederik~P Kingma and Max Welling.
\newblock Auto-encoding variational bayes, 2013.

\bibitem{luo2023latent}
Simian Luo, Yiqin Tan, Longbo Huang, Jian Li, and Hang Zhao.
\newblock Latent consistency models: Synthesizing high-resolution images with few-step inference, 2023.

\bibitem{fable2023showrunner}
Maas, Carey, Wheeler, Saatchi, Billington, and Shamash.
\newblock To infinity and beyond: Show-1 and showrunner agents in multi-agent simulations.
\newblock {\em arXiv preprint}, 2023.

\bibitem{mokady2022nulltext}
Ron Mokady, Amir Hertz, Kfir Aberman, Yael Pritch, and Daniel Cohen-Or.
\newblock Null-text inversion for editing real images using guided diffusion models, 2022.

\bibitem{X-T2003}
Chong-Wah Ngo, Ting-Chuen Pong, and Hong-Jiang Zhang.
\newblock Motion analysis and segmentation through spatio-temporal slices processing, 2003.

\bibitem{pham2023microscopic}
Sang~T. Pham, Natalia Koniuch, Emily Wynne, Andy Brown, and Sean~M. Collins.
\newblock Microscopic crystallographic analysis of dislocations in molecular crystals, 2023.

\bibitem{rombach2022highresolution}
Robin Rombach, Andreas Blattmann, Dominik Lorenz, Patrick Esser, and Björn Ommer.
\newblock High-resolution image synthesis with latent diffusion models, 2022.

\bibitem{gen3}
Runway.
\newblock Gen-3, 2024.

\bibitem{shykids2024}
Mike Seymour and Patrick Cederberg.
\newblock Actually using sora, 2024.

\bibitem{sohldickstein2015deep}
Jascha Sohl-Dickstein, Eric~A. Weiss, Niru Maheswaranathan, and Surya Ganguli.
\newblock Deep unsupervised learning using nonequilibrium thermodynamics, 2015.

\bibitem{song2022denoising}
Jiaming Song, Chenlin Meng, and Stefano Ermon.
\newblock Denoising diffusion implicit models, 2022.

\bibitem{supplementary}
Strikewind.
\newblock Supplementary materials: {https://strikewind.github.io/FYP-Supplementary/}, 2024.

\bibitem{gov_copyright}
Intellectual Property~Office UK.
\newblock Exceptions to copyright, 2021.

\bibitem{unterthiner2019accurate}
Thomas Unterthiner, Sjoerd van Steenkiste, Karol Kurach, Raphael Marinier, Marcin Michalski, and Sylvain Gelly.
\newblock Towards accurate generative models of video: A new metric \& challenges, 2019.

\bibitem{sdxllatent}
Timothy~Alexis Vass.
\newblock Explaining the sdxl latent space, 2024.

\bibitem{adventure_time}
Pendleton Ward, Frederator Studios, and Cartoon~Network Studios.
\newblock Adventure time, 2010.
\newblock TV Show.

\bibitem{zhang2023adding}
Lvmin Zhang, Anyi Rao, and Maneesh Agrawala.
\newblock Adding conditional control to text-to-image diffusion models, 2023.

\bibitem{opensora}
Zangwei Zheng, Xiangyu Peng, Shenggui Li, Hongxing Liu, and Yang You.
\newblock Open-sora, 2024.

\end{thebibliography}
}

\end{document}